\documentclass[10pt,twocolumn,letterpaper]{article}

\usepackage{cvpr}

\usepackage{times}
\usepackage{epsfig}
\usepackage{graphicx}
\usepackage{amsmath}
\usepackage{amssymb}
\usepackage{xspace}
\usepackage{subcaption}

\usepackage[breaklinks=true,bookmarks=false]{hyperref}

\makeatletter
\DeclareRobustCommand\onedot{\futurelet\@let@token\@onedot}
\def\@onedot{\ifx\@let@token.\else.\null\fi\xspace}

\def\etal{\emph{et al}\onedot}
\makeatother

\cvprfinalcopy 

\begin{document}\sloppy
	\title{Optical Flow-based 3D Human Motion Estimation from Monocular Video}
	
	\author{Thiemo Alldieck\qquad Marc Kassubeck\qquad Marcus Magnor 		\vspace{0.5em} 
		\\ Computer Graphics Lab, TU Braunschweig, Germany\\
		{\tt\small https://graphics.tu-bs.de}
	}
	
	\maketitle

\begin{abstract}
	We present a generative method to estimate 3D human motion and body shape from monocular video. Under the assumption that starting from an initial pose optical flow constrains subsequent human motion, we exploit flow to find temporally coherent human poses of a motion sequence. We estimate human motion by minimizing the difference between computed flow fields and the output of an artificial flow renderer. A single initialization step is required to estimate motion over multiple frames. Several regularization functions enhance robustness over time. Our test scenarios demonstrate that optical flow effectively regularizes the under-constrained problem of human shape and motion estimation from monocular video.
\end{abstract}
%
%
\section{Introduction}
\label{sec:intro}

Human pose estimation from video sequences has been an active field of research over the past decades and has various applications such as surveillance, medical diagnostics or human-computer interfaces \cite{moeslund2006survey}. A branch of human pose estimation is referred to as \textit{articulated motion parsing} \cite{zuffi2013estimating}. Articulated motion parsing defines the combination of monocular pose estimation and motion tracking in uncontrolled environments. This work presents a new approach to temporally coherent human shape and motion estimation in uncontrolled monocular video sequences. In contrast to earlier work, our work follows the \textit{generative} strategy, where both pose and shape parameters of a 3D body model are found to match the input image through analysis-by-synthesis \cite{magnor2015digital}.

For analysis of human motion, a 2D or 3D skeleton representation is often sufficient. On the other hand actual 3D geometry enables additional application scenarios. Especially image-based rendering applications can benefit from accurate 3D reconstruction. Examples are augmentation of an actor's garment \cite{rogge2014garment}, texture augmentation \cite{rav2008unwrap}, scene-space video processing \cite{klose2015sampling} or character deformation  \cite{jain2010moviereshape,zhou2010parametric}. Further applications can be found in virtual and augmented reality applications \cite{hauswiesner2013virtual}.

The 3D pose of a human figure is highly ambiguous when inferred from only a 2D image. Human silhouettes can often be explained by two or more poses \cite{guan2009estimating}. In this work we aim to limit the number of possible solutions by taking the projected movement of the body, namely optical flow, into account. Besides the motion of individual body parts, optical flow contains information about boundaries of rigid structures. Additionally optical flow is an abstraction layer to the input modality. Unique appearance effects such as texture and shading are removed \cite{fablet2002automatic,romero2015flowcap}. These features make optical flow highly suitable for generative optimization problems.



\begin{figure}
	\centering
	\begin{subfigure}[b]{0.33\columnwidth}
		\includegraphics[width=\textwidth]{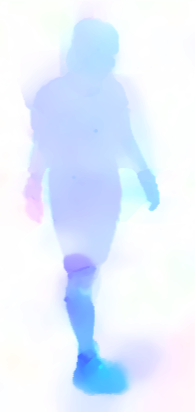}
	\end{subfigure}%
	~
	\begin{subfigure}[b]{0.33\columnwidth}
		\includegraphics[width=\textwidth]{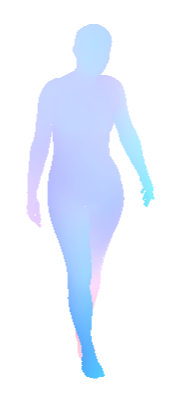}
	\end{subfigure}%
	~
	\begin{subfigure}[b]{0.33\columnwidth}
		\includegraphics[width=\textwidth]{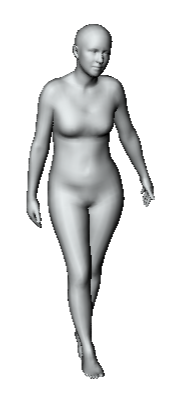}
	\end{subfigure}
	\caption{Method overview. Color-coded observed flow, estimated flow and resulting pose.}
	\label{fig:observedvsoptimized}
\end{figure}

Existing methods for generative 3D human motion estimation from monocular video find a pose per frame independently. These poses are not necessarily related and can result in significant jumps in position and joint angles between two successive frames. Our work aims for a more global solution to the problem, that ensures that the estimated poses are temporal coherent and therefore form fluent human motion. One of the key contributions of this work is the formulation of an artificial flow renderer that is used in the optimization process. The renderer calculates pixel displacement between the current scene and a previous time step. Having a flow renderer available, human motion estimation can be formulated as an analysis-by-synthesis problem.
The main idea of this work is that the current pose of a subject is well predetermined by the pose of the last frame and the optical flow between the last and current frame. Following this idea, we search for those model parameters that create the highest similarity between observed and rendered flow for the two input frames. 
High temporal consistency between the reconstructed poses is achieved by initializing the current pose based on a Kalman filter prediction and optimization on the optical flow to the last pose.

We evaluate the proposed method using two well known datasets. We show the temporal consistency of our approach qualitatively and evaluate its 3D and 2D precision quantitatively. In the first test, we compare 3D joint positions against ground truth of the HumanEva-I dataset \cite{sigal2010humaneva} and results of two recently published methods \cite{bogo2016smplify,wandt20163d}. The second evaluation compares projected 2D joint positions against ground truth of the VideoPose 2.0 dataset \cite{sapp2011parsing} featuring camera movement and rapid gesticulation. We compare our results against  a recent method for joint localization \cite{deepcut16cvpr}. Results demonstrate the strengths and potential of the proposed method.


\section{Related Work}
Human pose estimation is a broad and active field of research. We focus on 3D model-based approaches and previous work that exploits optical flow during pose estimation.

\textbf{Human pose from images.}
3D human pose estimation is often based on the use of a body model. Human body representations exist in 2D and 3D. Many of the following methods utilize the 3D human body model SCAPE \cite{anguelov2005scape}. SCAPE is a deformable mesh model learned from body scans. Pose and shape of the model are parametrized by a set of body part rotations and low dimensional shape deformations. 
In recent work the SMPL model, a more accurate blend shape model compatible with existing rendering engines, has been presented by Loper \etal \cite{smpl2015loper}.

A variety of approaches to 3D pose estimation have been presented using various cues including shape from shading, silhouettes and edges. Due to the highly ill-posed and under-constrained nature of the problem these methods often require user interaction e.g.\ through manual annotation of body joints on the image \cite{taylor2000reconstruction,parameswaran2004view}.

Guan \etal \cite{guan2009estimating} have been the first to present a detailed method to recover human pose together with an accurate shape estimate from single images. Based on manual initialization, parameters of the SCAPE model are optimized exploiting edge overlap and shading. The work is based on \cite{bualan2007detailed}, a method that recovers the 3D pose from silhouettes from 3-4 calibrated cameras. Similar methods have been presented by B{\u{a}}lan \etal \cite{bualan2007shining} and Sigal \etal \cite{sigal2007combined}, also requiring multi-view input.
Hasler \etal \cite{hasler2010multilinear} fit a statistical body model \cite{hasler2009statistical} into monocular image silhouettes. A similar approach is followed by Chen \etal \cite{chen2010inferring}. In recent work, Bogo \etal.  \cite{bogo2016smplify} present the first method to extract both pose and shape from a single image fully automatically. 2D joint locations are found using the CNN-based approach DeepCut \cite{deepcut16cvpr}, then projected joints of the SMPL model are fitted against the 2D locations. The presented method is similar to ours as it also relies on 2D features. In contrast to our work no consistency with the image silhouette or temporal coherency is guaranteed.

\textbf{Pose reconstruction for image based rendering.} 
3D human pose estimation can serve as a preliminary step for image based rendering techniques. In early work Carranza  \etal \cite{carranza2003free} have been the first to present free-viewpoint video using model-based reconstruction of human motion using the subject's silhouette in multiple camera views. Zhou \etal \cite{zhou2010parametric} and Jain \etal \cite{jain2010moviereshape} present updates to model-based pose estimation for subsequent reshaping of humans in images and videos respectively. Rogge \etal \cite{rogge2014garment} fit a 3D model for automatic cloth exchange in videos. All methods utilize various cues, none of them uses optical flow for motion estimation.

\textbf{Optical flow based methods.}
Different works have been presented exploiting optical flow for different purposes. Sapp \etal \cite{sapp2011parsing} and Fragkiadaki \etal \cite{fragkiadaki2013pose} use optical flow for segmentation as a preliminary step for pose estimation. Both exploit the rigid structure revealing property of optical flow, rather than information about motion. Fablet and Black \cite{fablet2002automatic} use optical flow to learn motion models for automatic detection of human motion. Efros \etal \cite{efros2003recognizing} categorize human motion viewed from a distance by building an optical flow-based motion descriptor. Both methods label motion without revealing the underlying movement pattern.

 In recent work, Romero \etal \cite{romero2015flowcap} present a method for 2D human pose estimation using optical flow only. They detect body parts by porting the random forest approach used by the Microsoft Kinect to use optical flow. Brox \etal \cite{brox2006high} have shown that optical flow can be used for 3D pose tracking of rigid objects. They propose the use for objects \emph{modeled as kinematic chains}. They argue that optical flow provides point correspondences inside the object contour which can help to identify a pose where silhouettes are ambiguous. Inspired by the above mentioned characteristics, this paper investigates as to what extent optical flow can be used for 3D human motion estimation from monocular video.

 \section{Method}
 
 Optical flow \cite{gibson1950perception} is the perception of motion by our visual sense. For two successive video frames, it is described as a 2D vector field that matches a point in the first frame to the displaced point in the following frame \cite{horn1981determining}. 
 Although calculated in the image plane, optical flow contains 3D information, as it can be interpreted as the projection of 3D scene flow \cite{vedula1999three}. On the other hand, optical flow is, by definition, not independent of lighting changes and highly dependent on high-contrast edges. 
 In this work we assume that all observed surfaces are diffuse, opaque and textured. Under this assumption, the entire observed optical flow is caused by relative movement between object and camera.
 
 The presented method estimates pose parameters and position of a human model frame by frame. The procedure requires a single initialization step and then runs automatically. The parameters for each frame are found by moving the human model along the optical flow field between the current frame and the subsequent frames. A set of regularization functions is defined to keep the accumulated error over time to a minimum to make the method robust. The modular structure of the presented method allows for further optimization.
 
 In the following we elaborate on the used methods and objectives. Afterwards the initialization and optimization procedures are described.
 
 \subsection{Scene Model}
 In this work, we use the human body model SMPL \cite{smpl2015loper}. 
 The model can be reshaped using 10 shape parameters $\vec{\beta}$. For different poses, 72 pose parameters $\vec{\theta}$ can be set, including global orientation. $\vec{\beta}$ and $\vec{\theta}$ produce realistic vertex transformations and cover a large range of body shapes and poses. 
 
 We define $(\gamma, \vec{\beta}, \vec{\theta}_i, \vec{\sigma}_i)$ as the model state at time step $i$, with global translation vector $\vec{\sigma}$ and gender $\gamma$. 
 
 We assume that the camera position and rotation as well as its focal length are known and static. It is however not required that the camera of the actual scene is fixed, as the body model can rotate and move around the camera.

 \subsection{Flow Renderer}
 
 The core of the presented method is our differential flow renderer built upon OpenDR \cite{loper2014opendr}, a powerful open source framework for analysis-by-synthesis. The rendered flow image depends on the vertex locations determined by the virtual human model's pose parameters $\vec{\theta}$ and its translation $\vec{\sigma}$. To be able to render the flow \emph{in situ}, we calculate the flow from frame $i$ to $i-1$, referred as backward flow. With this approach each pixel, and more importantly, each vertex location contains the information where it came from rather than were it went and can be rendered in place.
 
 The calculation of the flow is achieved as follows: The first step calculates the displacement of all vertices between two frames $i$ and $j$ in the image plane.
 Then the flow per pixel is calculated through barycentric interpolation of the neighboring vertices. Visibility and barycentric coordinates are calculated through the standard OpenGL rendering pipeline. 

The core feature of the utilized rendering framework OpenDR is the differentiability of the rendering pipeline. To benefit from that property, our renderer estimates the partial derivatives of each flow vector with respect to each projected vertex position. 

 \begin{figure*}
 	\centering
 	\begin{subfigure}[b]{0.2\textwidth}
 		\includegraphics[width=\textwidth]{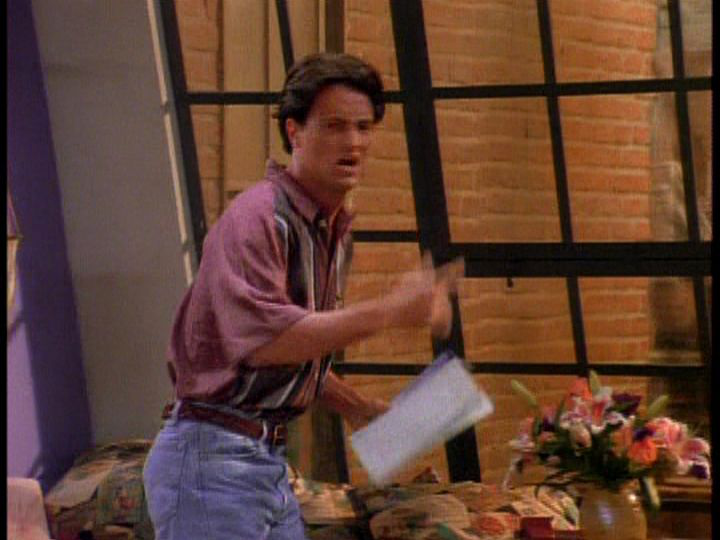}
 	\end{subfigure}%
 	\begin{subfigure}[b]{0.2\textwidth}
 		\includegraphics[width=\textwidth]{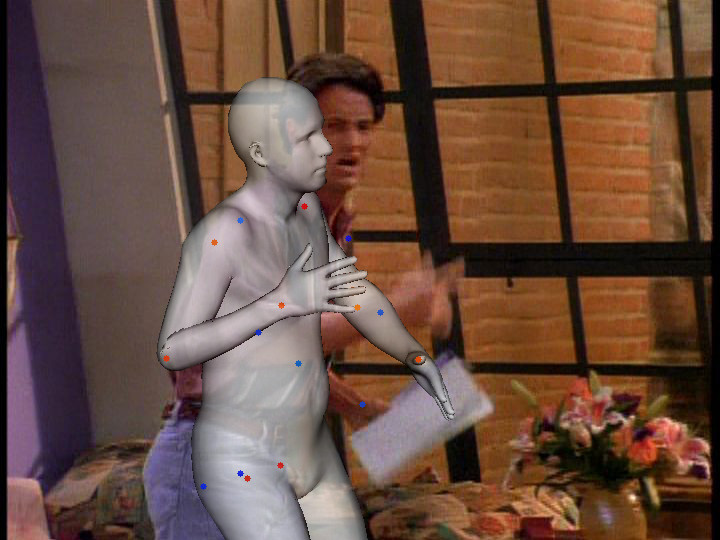}
 	\end{subfigure}%
 	\begin{subfigure}[b]{0.2\textwidth}
 		\includegraphics[width=\textwidth]{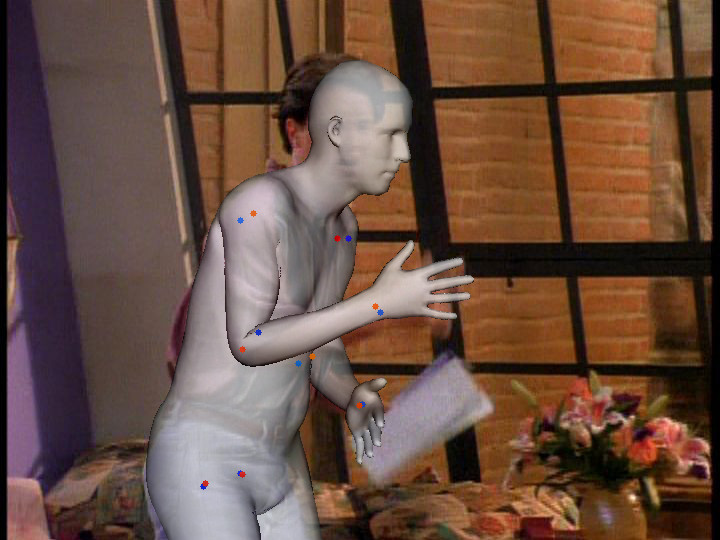}
 	\end{subfigure}%
 	\begin{subfigure}[b]{0.2\textwidth}
 		\includegraphics[width=\textwidth]{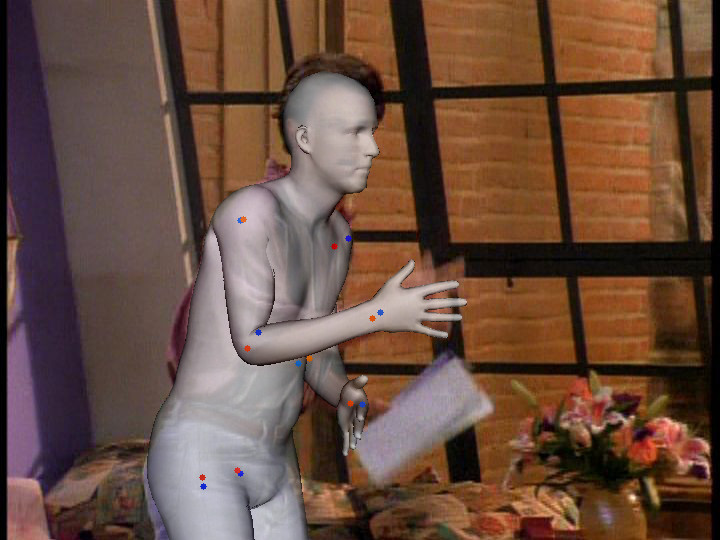}
 	\end{subfigure}%
 	\begin{subfigure}[b]{0.2\textwidth}
 		\includegraphics[width=\textwidth]{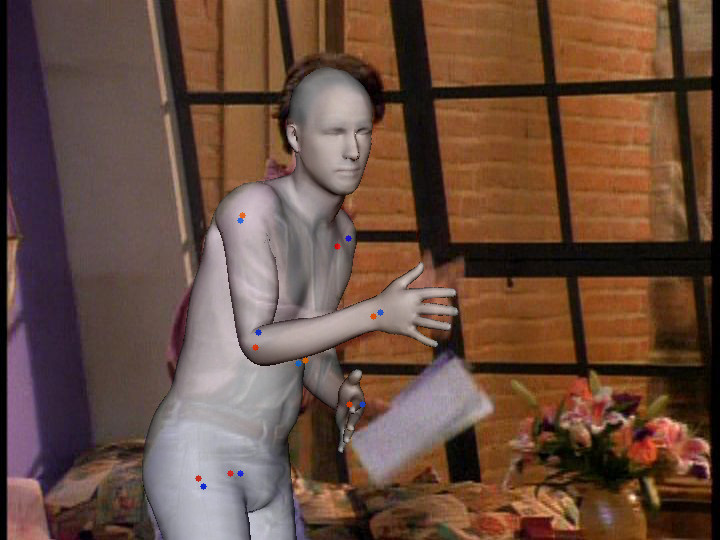}
 	\end{subfigure}
 	\caption{Method initialization. Observed image, manual pose initialization, first optimization based on joint positions (red: model joints; blue: manually marked joints), second optimization including silhouette coverage, optical flow based correction.}
 	\label{fig:init}
 \end{figure*}

\subsection{Flow Matching}

Having a flow renderer available, we can formulate the pose estimation as an optimization problem. The cost function $E_f$ over all pixels $p$ is defined as follows:
\begin{equation}
E_f = \sum_{p}|| F_{\text{o}}(i, i-1, p) - F_{\text{r}}(i, i-1, p) ||^2
\end{equation}
\noindent where $F_{\text{r}}$ refers to the \emph{rendered} and $F_{\text{o}}$ to the \emph{observed} flow field calculated on the input frames $i$ and $i-1$. The objective drives the optimization in such way that the rendered flow is similar to the observed flow (Fig. \ref{fig:observedvsoptimized}). As proposed in \cite{loper2014opendr}, we evaluate $E_f$ not over the flow field but over its Gaussian pyramid in order to perform a more global search.

For this work we use the method by Xu \etal \cite{xu2012motion} to calculate the observed optical flow field. The method has its strength in the ability to calculate large displacements while at the same time preserving motion details and handling occlusions.

The definition of the objective shows that the performance of the optical flow estimation is crucial to the overall performance of the presented method. To compensate for inaccuracies of the flow estimation and to lower the accumulated error over time, we do not rely exclusively on the flow for pose estimation, but employ regularization as well.

\subsection{Pose Prior}

SMPL does not define bounds for deformation. We introduce soft boundaries to constrain the joint angles in form of a cost function for pose estimation:
\begin{equation}
E_b = || \max(e^{\vec{\theta}_{\text{min}} - \vec{\theta}_i} - 1, 0) + \max(e^{\vec{\theta}_i - \vec{\theta}_{\text{max}}} - 1, 0) ||^2
\end{equation}
\noindent
where $\vec{\theta}_{\text{min}}$ and $\vec{\theta}_{\text{max}}$ are empirical lower and upper boundaries and $e$ and $\max$ are applied component-wise.

Furthermore, we introduce extended Kalman filtering per joint and linear Kalman filtering for translation. Besides for temporal smoothness the Kalman filters are used to predict an \emph{a priori} pose for the next frame before optimization.

During optimization the extremities of the model may intersect with other body parts. To prevent this, we integrate the interpenetration error term $E_{sp}$ from \cite{bogo2016smplify}. The error term is defined over a capsule approximation of the body model. By using an error term interpenetration is not strictly prohibited but penalized.

\subsection{Silhouette Coverage}


Pose estimation based on flow similarity requires that the rendered human model accurately covers the subject in the input image. Only body parts that cover the correct counterpart in the image can be moved correctly based on flow. To address inaccuracies caused by flow calculation, we introduce boundary matching.

We use the method presented by B{\u{a}}lan \etal \cite{bualan2007detailed} and adapt it to achieve differentiability. A cost function measures how well the model fits the image silhouette $S_I$ by penalizing non-overlapping pixels by the shortest distance to the model silhouette $S_M$. For this purpose Chamfer distance maps $C_I$ for the image silhouette  and $C_M$ for the model are calculated. 
The cost function is defined as:

\begin{equation}
E_c = \sum_{p} || a S_{M_i}(p)C_I(p) + (1-a)S_I(p)C_{M_i}(p)||^2
\end{equation}

\noindent where $a$ weighs $S_{M_i}C_I$ stronger as image silhouettes are wider so that it is more important for the model to reside within in the image silhouette than to completely cover it. To achieve differentiability we approximate $C_M$ by calculating the shortest distance of each pixel to the model capsule approximation. To lower computation time, we calculate only a grid of values and interpolate in between.


\subsection{Initialization}

For the initialization of the presented method two manual steps are required. First the user sets the joints of the body model to a pose that roughly matches the observed pose. It is sufficient that only the main joints such as shoulder, elbow, hip and knee are manipulated. In a second step the user marks joint locations of hips, knees, ankles, shoulders, elbows and wrists in the first frame. If the position of a joint cannot be seen or estimated it may be skipped. From this point no further user input is needed.

The initialization is then performed in three steps (Fig. \ref{fig:init}). The first step minimizes the distance between the marked joints and their model counterparts projected to the image plane, while keeping $E_{sp}$ and $E_b$ low. We optimize over translation $\vec{\sigma}$, pose $\vec{\theta}$ and shape $\vec{\beta}$. To guide the process we regularize both $\vec{\theta}$ and $\vec{\beta}$ with objectives that penalize high differences to the manually set pose and the mean shape. In the second step we include the silhouette coverage objective $E_c$. Finally, we optimize the estimated pose for temporal consistency. We initialize the second frame with the intermediate initialization result and optimize on the flow field afterwards. While optimizing $E_f$ we still allow updates for $\vec{\theta}_0$ and $\vec{\sigma}_0$.







\subsection{Optimization}

After initialization we now iteratively find each pose using the defined objectives. Each frame is initialized with the Kalman prediction step. Afterwards we minimize:
\begin{equation}
\begin{aligned} 
\min_{\vec{\sigma}, \vec{\theta}} (\lambda_{f} E_f + \lambda_{c} E_c + \lambda_{b} E_b + \lambda_{sp} E_{sp} + \lambda_{\theta} E_\theta)
\end{aligned}
\end{equation}
with scalar weights $\lambda$. $E_\theta$ regularizes the current pose with respect to the last pose.

\section{Evaluation}

\begin{figure*}
	\centering
	\begin{subfigure}[b]{0.20\textwidth}
		\includegraphics[width=\textwidth]{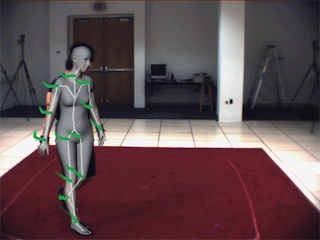}
	\end{subfigure}%
	\begin{subfigure}[b]{0.20\textwidth}
		\includegraphics[width=\textwidth]{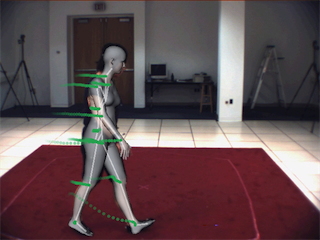}
	\end{subfigure}%
	\begin{subfigure}[b]{0.20\textwidth}
		\includegraphics[width=\textwidth]{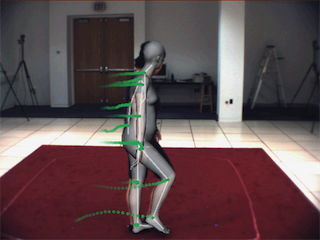}
	\end{subfigure}%
	\begin{subfigure}[b]{0.20\textwidth}
		\includegraphics[width=\textwidth]{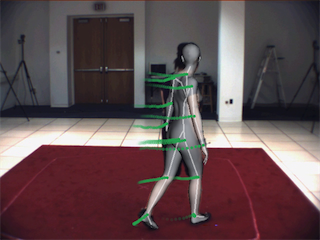}
	\end{subfigure}%
	\begin{subfigure}[b]{0.20\textwidth}
		\includegraphics[width=\textwidth]{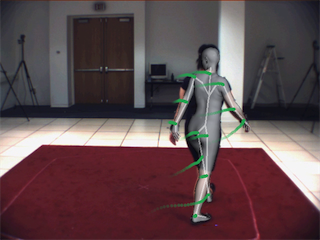}
	\end{subfigure}
	\begin{subfigure}[b]{0.20\textwidth}
		\includegraphics[width=\textwidth]{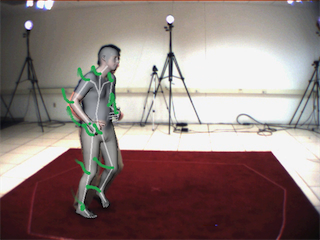}
	\end{subfigure}%
	\begin{subfigure}[b]{0.20\textwidth}
		\includegraphics[width=\textwidth]{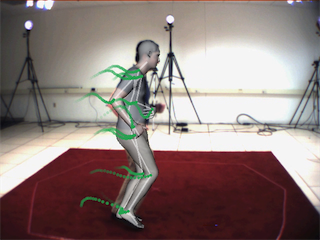}
	\end{subfigure}%
	\begin{subfigure}[b]{0.20\textwidth}
		\includegraphics[width=\textwidth]{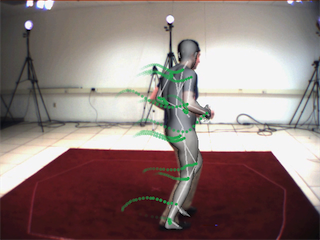}
	\end{subfigure}%
	\begin{subfigure}[b]{0.20\textwidth}
		\includegraphics[width=\textwidth]{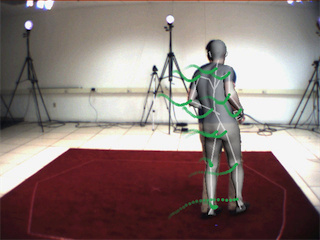}
	\end{subfigure}%
	\begin{subfigure}[b]{0.20\textwidth}
		\includegraphics[width=\textwidth]{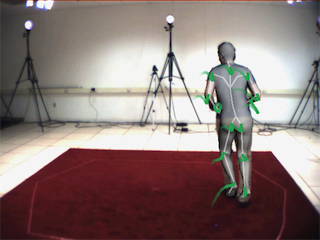}
	\end{subfigure}
	\caption{Resultant poses of frames 25 to 125 of the HumanEva-I test sets. Green traces show the history of evaluated joints.}
	\label{fig:results_he}
\end{figure*}

We evaluate the 3D and 2D pose accuracy of the presented method using two publicly available datasets: HumanEva-I \cite{sigal2010humaneva} and VideoPose2.0 \cite{sapp2011parsing}. Ground truth is available for both datasets. We compare our results in both tests, 3D and 2D, against state-of-the-art methods \cite{bogo2016smplify,wandt20163d,deepcut16cvpr}. Foreground masks needed for our method have been hand-annotated using an open-source tool for image annotation\footnote {https://bitbucket.org/aauvap/multimodal-pixel-annotator}.

\textbf{HumanEva-I.} The HumanEva-I datasets features different actions performed by 4 subjects filmed under laboratory conditions. We reconstruct 130 frames of the sets \emph{Walking C1} by subject 1 and \emph{Jog C2} by subject 2 without reinitialization. The camera focal length is known. We do not adjust our method for the dataset except setting the $\lambda$ weights.

Fig. \ref{fig:results_he} shows a qualitative analysis. The green plots show the history of the joints used for evaluation. The traces demonstrate clearly the temporal coherence of the presented method. The low visual error in the last frames demonstrates  that the presented method is robust over time.

We compare our method against the recent methods of Bogo \etal. \cite{bogo2016smplify} and Wandt \etal \cite{wandt20163d}. We use \cite{bogo2016smplify} without the linear pose regressor learned for the HumanEva sequences, which is missing in the publicly available source code. Frames that could not be reconstructed because of undetected joints have been excluded for evaluation. The 3D reconstruction of \cite{wandt20163d} is initialized with the same DeepCut \cite{deepcut16cvpr} results as used for \cite{bogo2016smplify}. 

We measure the precision of the methods by calculating the euclidean distance of 13 3D joint locations to ground truth locations from MoCap data. Beforehand, we achieve the optimal linear alignment of the results of all methods by Procrustes analysis. In order to demonstrate the global approach of our method, we follow two strategies here: First we measure the joint error after performing Procrustes per frame. Afterwards we calculate a per sequence alignment over all joint locations in all frames and measure the resulting mean error. Table \ref{tab:res3d} shows the result of all tests.

The results show that our method performs best in three of four test scenarios. In contrast to the methods of Bogo \etal. \cite{bogo2016smplify} and Wandt \etal \cite{wandt20163d}, our method does not require prior knowledge about the performed motion or a trained pose prior. We explain the performance with the fact, that each frame is well pre-initialized by its predecessor and a robust Kalman prediction. This strength is especially noticeable in the global analysis. Both compared methods can give no guarantee about the 3D position of the estimated pose, which may result in jumps especially along the Z-axis.

\textbf{VideoPose2.0.} After evaluation with fixed camera and under laboratory conditions, we test our method under a more challenging setting. The second evaluation consists of three clips of the VideoPose2.0 dataset. We choose the "fullframe, every frame" ($720 \times 540$px) variant in order to face camera movement. Ground truth is given in form of projected 2D location of shoulders, elbows and wrists for every other frame. The camera focal length has been estimated.

We evaluate our method in 2D by comparison against DeepCut \cite{deepcut16cvpr}, the same method that has been used before as input for the 3D reconstruction methods. 
Table \ref{tab:res2d} shows the mean euclidean distance to ground truth 2D joint locations. We use the first detected person by DeepCut and exclude several undetected joints from its evaluation. For our method, we project the reconstructed 3D joint locations to the image plane. The mixed performance of \cite{deepcut16cvpr} is due to problems of the CNN with background objects. The results show that our method produces similar precision while providing much more information. However, the increasing performance of CNN-based methods suggests that our method can benefit from semantic scene information for reinitialization in future work.

\begin{table}[t]
	\begin{center}
		\caption{Mean 3D joint error in cm for local per frame and global per sequence Procrustes analysis.}
		\label{tab:res3d}
		\begin{tabular}{ccccc}
			\hline 
			& \multicolumn{2}{c}{Walking S1 C1} & \multicolumn{2}{c}{Jog S2 C2}  \\ 
			& local & global & local & global \\ 
			\hline 
			Bogo \etal \cite{bogo2016smplify} & 6.6  & 17.4 & 7.5 & 10.4  \\ 
			Wandt \etal \cite{wandt20163d} & 5.7 & 34.0 & \textbf{6.3} & 38.0 \\ 
			Our method & \textbf{5.5}  & \textbf{7.6} & 7.9 & \textbf{9.9} \\
			\hline 
		\end{tabular}
	\end{center}
\end{table}

\begin{table}[t]
	\begin{center}
		\caption{Mean 2D joint error in pixels.}
		\label{tab:res2d}
		\begin{tabular}{ccccc}
			\hline 
			& Chandler & Ross & Rachel  \\ 
			\hline 
			DeepCut \cite{deepcut16cvpr} &  25.3 & \textbf{10.5}  & 32.8  \\ 
			Our method & \textbf{23.3} & 21.9 & \textbf{15.9}  \\
			\hline 
		\end{tabular}
	\end{center}
\end{table}

\begin{figure}
	\centering
	\begin{subfigure}[b]{0.33\linewidth}
		\includegraphics[width=\linewidth]{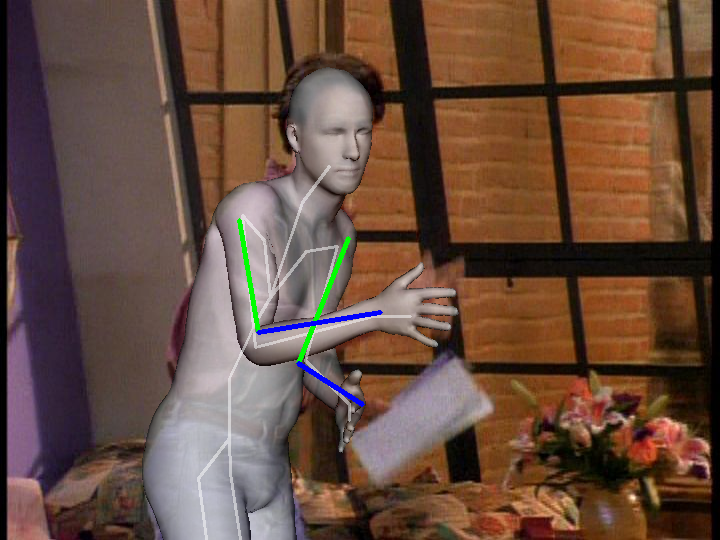}
	\end{subfigure}%
	\begin{subfigure}[b]{0.33\linewidth}
		\includegraphics[width=\linewidth]{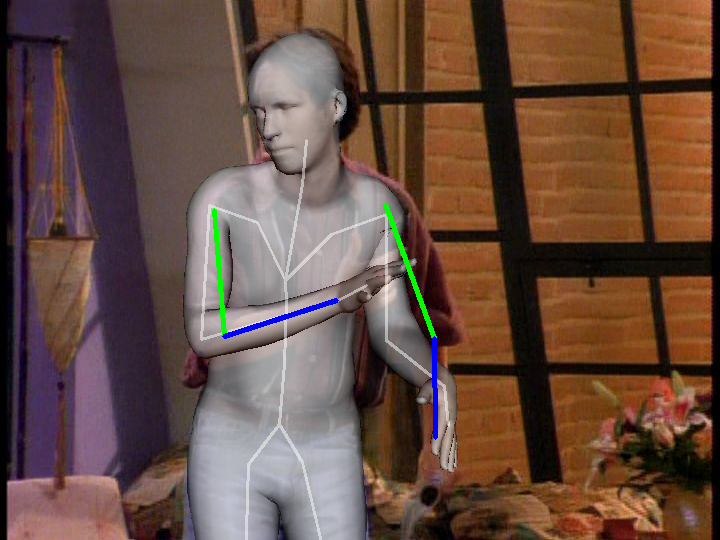}
	\end{subfigure}%
	\begin{subfigure}[b]{0.33\linewidth}
		\includegraphics[width=\linewidth]{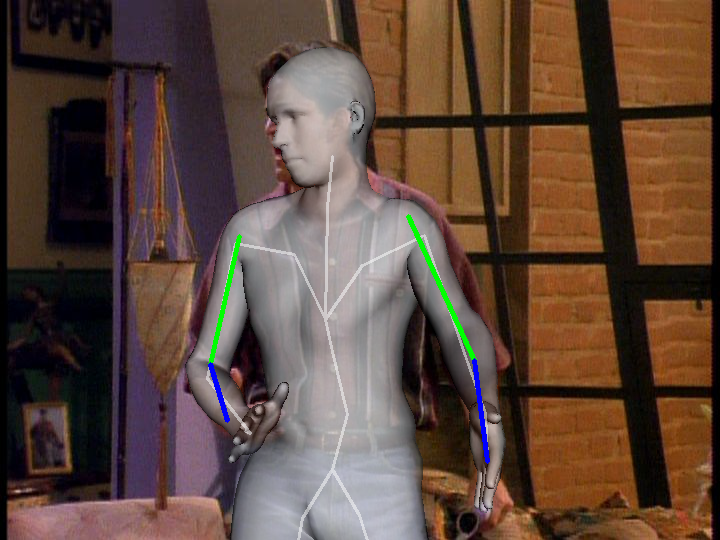}
	\end{subfigure}
	\begin{subfigure}[b]{0.33\linewidth}
		\includegraphics[width=\linewidth]{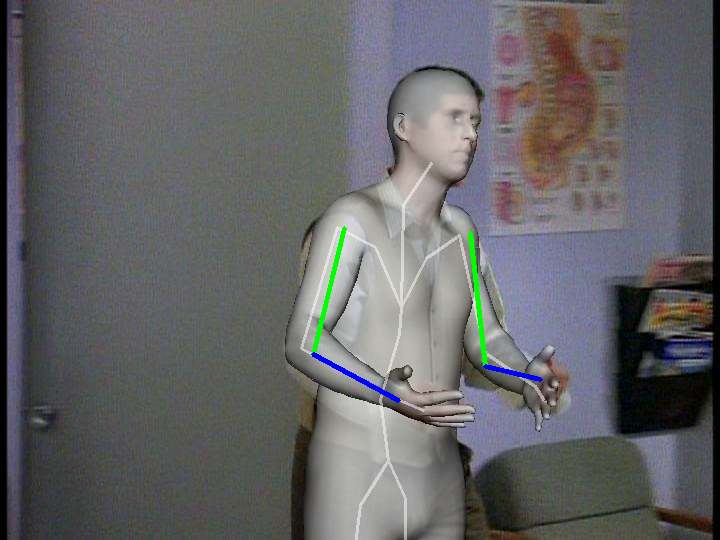}
	\end{subfigure}%
	\begin{subfigure}[b]{0.33\linewidth}
		\includegraphics[width=\linewidth]{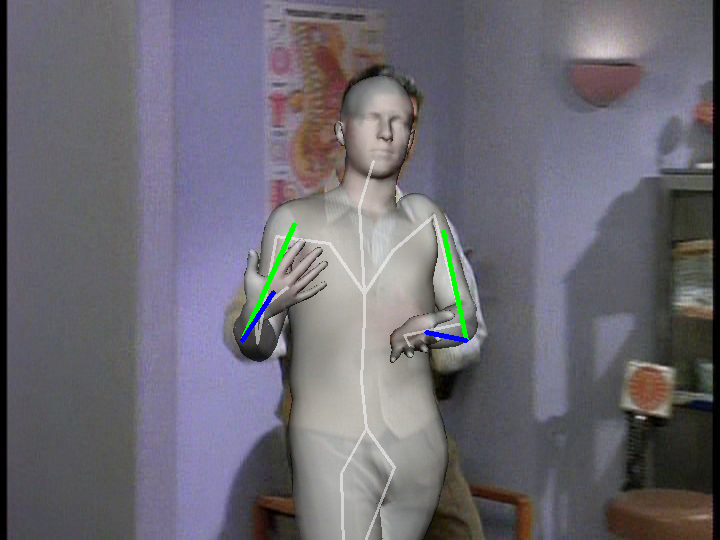}
	\end{subfigure}%
	\begin{subfigure}[b]{0.33\linewidth}
		\includegraphics[width=\linewidth]{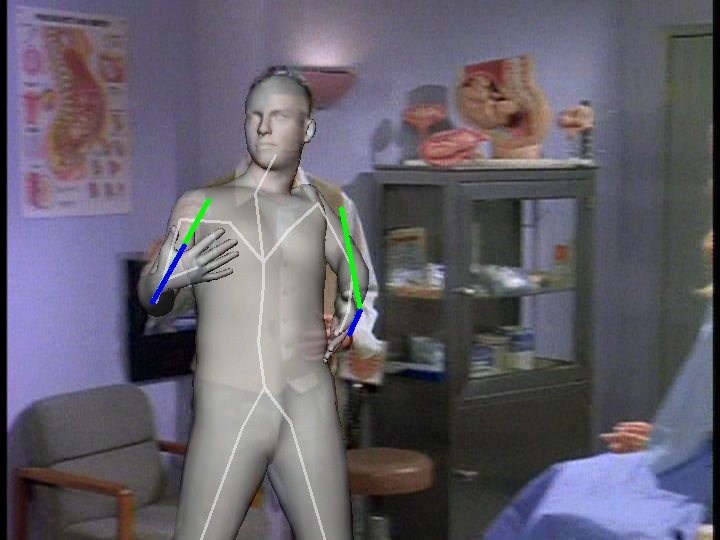}
	\end{subfigure}
	\begin{subfigure}[b]{0.33\linewidth}
		\includegraphics[width=\linewidth]{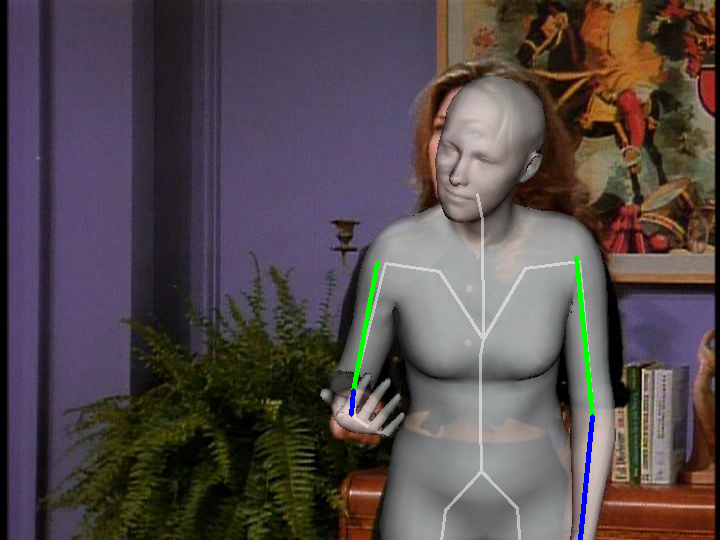}
	\end{subfigure}%
	\begin{subfigure}[b]{0.33\linewidth}
		\includegraphics[width=\linewidth]{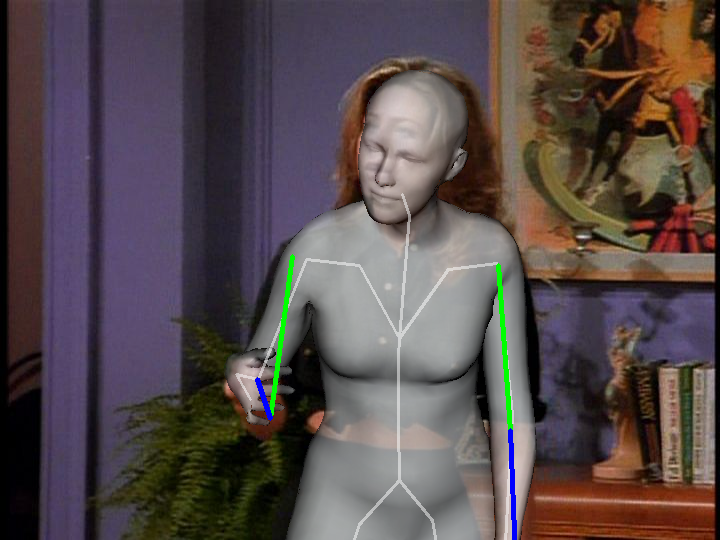}
	\end{subfigure}%
	\begin{subfigure}[b]{0.33\linewidth}
		\includegraphics[width=\linewidth]{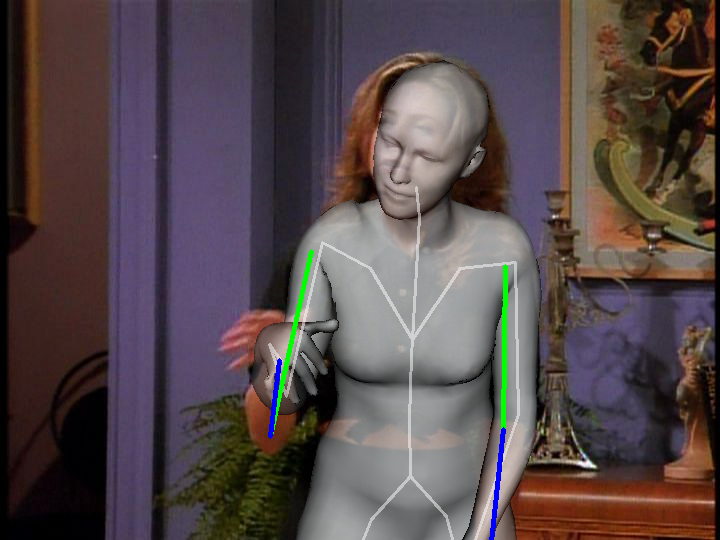}
	\end{subfigure}
	\caption{Resultant poses of frames 1, 21 and 41 of the VideoPose2.0 sets (Chandler, Ross, Rachel) with ground truth arm locations (green and blue).}
	\label{fig:results_friends}
\end{figure}

\section{Conclusions}	

We have presented a new method for estimating 3D human motion from monocular video footage. The approach utilizes optical flow to recover human motion over time from a single initialization frame. For this purpose a novel flow renderer has been developed that enables direct interpretation of optical flow. The rich human body model SMPL provides the description of estimated human motion. Different test cases have shown the applicability of the approach.

Our work is focused on automatic estimation of human motion from monocular video. In future work we plan to further automatize our method and increase robustness over longer periods. The method might benefit from recent developments in semantic segmentation \cite{oliveira2016deep} and human joint angle limits \cite{akhter2015poseconditioned}. Building upon the presented framework, the next steps are texturing of the model and geometry refinement, enabling new video editing and virtual reality applications.

\hspace{2mm}

\noindent
\begin{minipage}{\linewidth}
\textbf{Acknowledgments}\\
The authors gratefully acknowledge funding by the German Science Foundation from project DFG MA2555/4-1.
\end{minipage}


\end{document}